\documentclass{article}
\usepackage{spconf,amsmath,graphicx}
\usepackage{epstopdf}
\usepackage{lipsum}
\usepackage{todonotes}
\usepackage{textcomp}
\usepackage{xspace}
\usepackage[perc]{overpic}
\usepackage{color}
\usepackage[absolute]{textpos}
\usepackage[bottom]{footmisc}

\newcommand{\figNum}[2]{\textcolor{#1}{\fontfamily{pnc}\selectfont\textbf{\small #2}}}

\newcommand{\CARE}{\mbox{\textsc{CARE}}\xspace}
\newcommand{\NoiseNoise}{\mbox{\textsc{Noise2Noise}}\xspace}
\newcommand{\ProjectionProjection}{\mbox{\textsc{Projection2Projection}}\xspace}
\newcommand{\TomoTomo}{\mbox{\textsc{Tomo2Tomo}}\xspace}
\newcommand{\PtoP}{\mbox{\textsc{P2P}}\xspace}
\newcommand{\TtoT}{\mbox{\textsc{T2T}}\xspace}
\newcommand{\TOMO}{\mbox{\textsc{Tomo110}}\xspace}
\newcommand{\EMPIAR}{\mbox{\textsc{EMPIAR-10110}}\xspace}

\newcommand{\compactsubsub}[2]{\vspace{1mm}\textit{#1:} #2}

\makeatletter
\DeclareRobustCommand\onedot{\futurelet\@let@token\@onedot}
\def\@onedot{\ifx\@let@token.\else.\null\fi\xspace}
\def\eg{\emph{e.g}\onedot} 
\def\ie{\emph{i.e}\onedot}

\makeatother

\title{CRYO-CARE: CONTENT-AWARE IMAGE RESTORATION\\ FOR CRYO-TRANSMISSION ELECTRON MICROSCOPY DATA$^*$}

\name{Tim-Oliver Buchholz$^{1,2}$, Mareike Jordan$^{2}$, Gaia Pigino$^{2}$, Florian Jug$^{1,2}$}
\address{
$^{1}$Center for Systems Biology Dresden (CSBD)\\
$^{2}$Max-Planck Institute of Molecular Cell Biology and Genetics}

\begin{document}
\maketitle
%
\newcommand\figPtoP{
\begin{figure}[hbt]
\centerline{
\begin{minipage}{\linewidth}
\begin{overpic}[width=2.8cm]{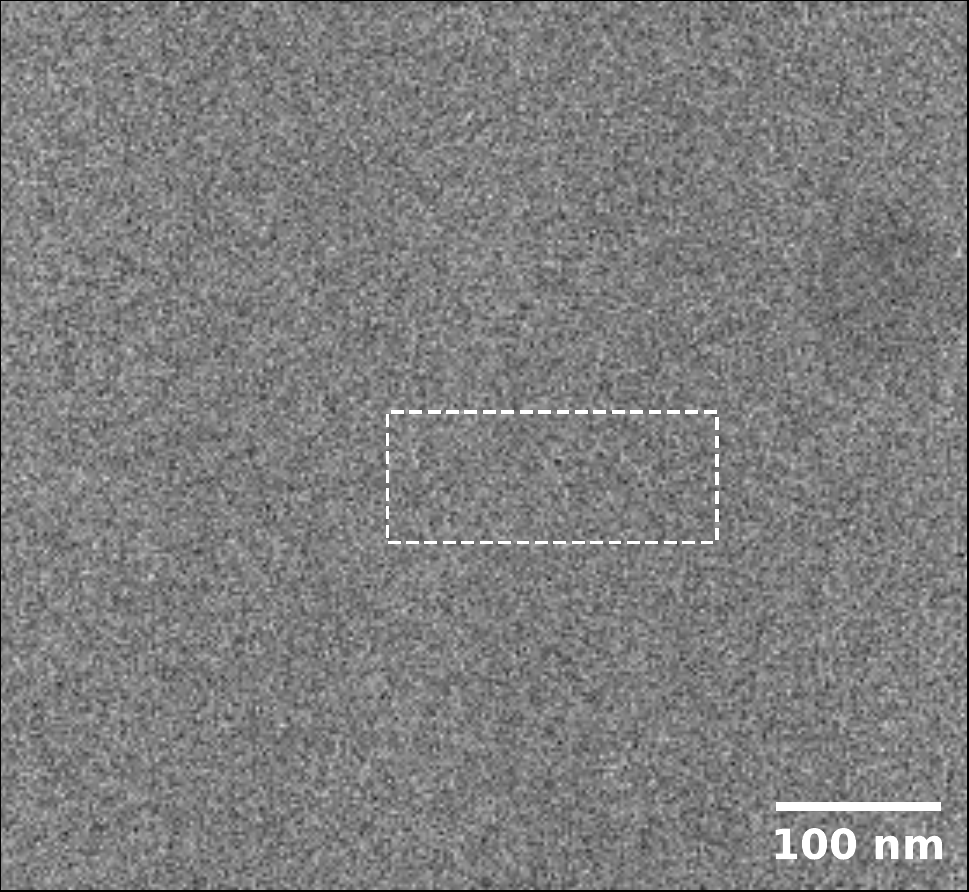}
 \put (0.5,2.75) {\figNum{white}{(a)}}
\end{overpic}
\begin{overpic}[width=2.8cm]{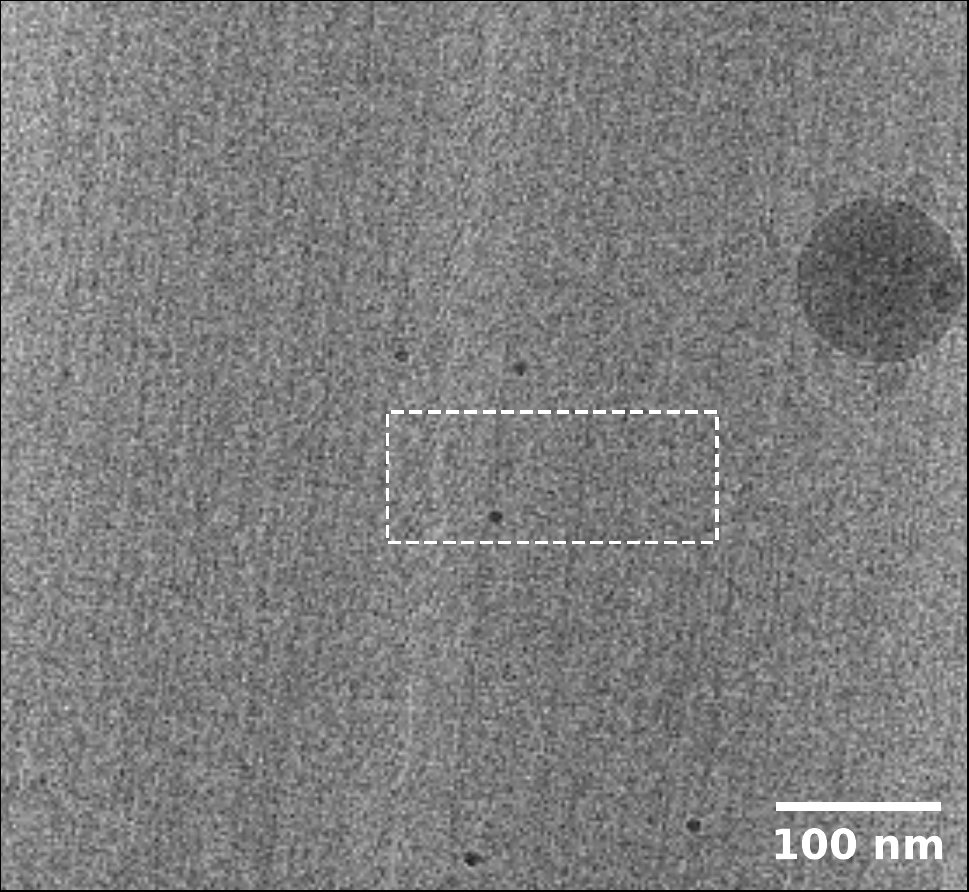}
 \put (0.5,2.75) {\figNum{white}{(b)}}
\end{overpic}
\begin{overpic}[width=2.8cm]{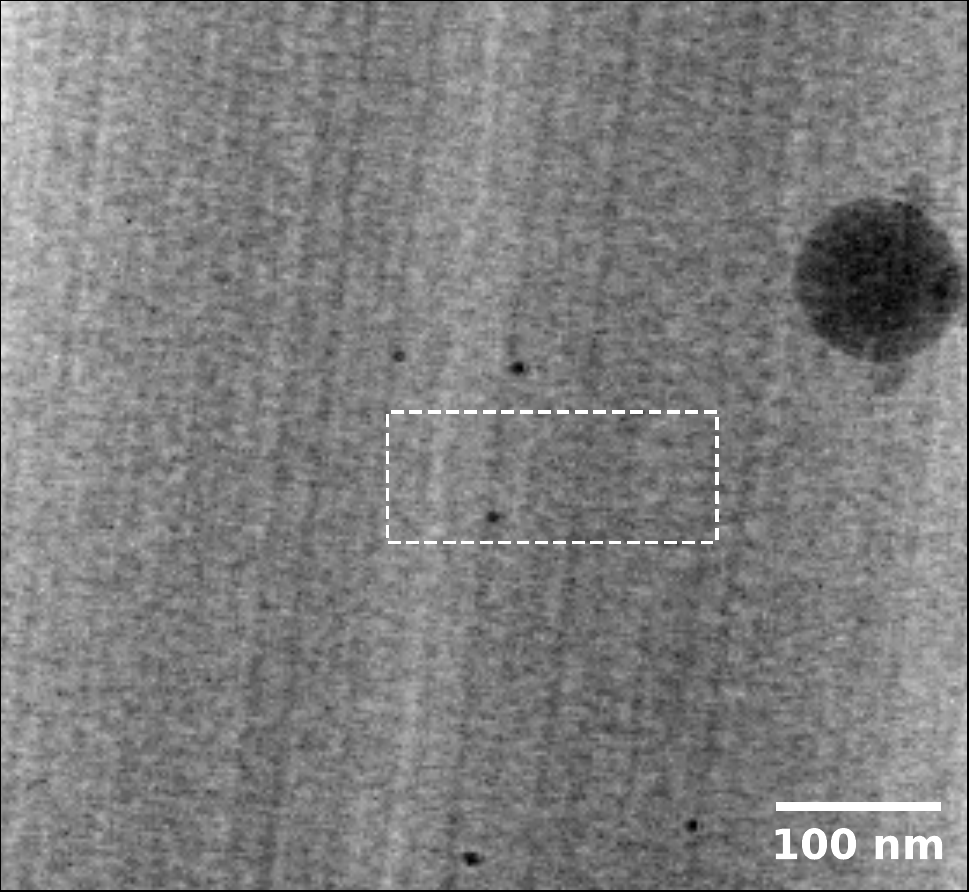}
 \put (0.5,2.75) {\figNum{white}{(c) P2P-\textit{tap}}}
\end{overpic}
\end{minipage}
}
\centerline{
\begin{minipage}{\linewidth}
\includegraphics[width=2.8cm]{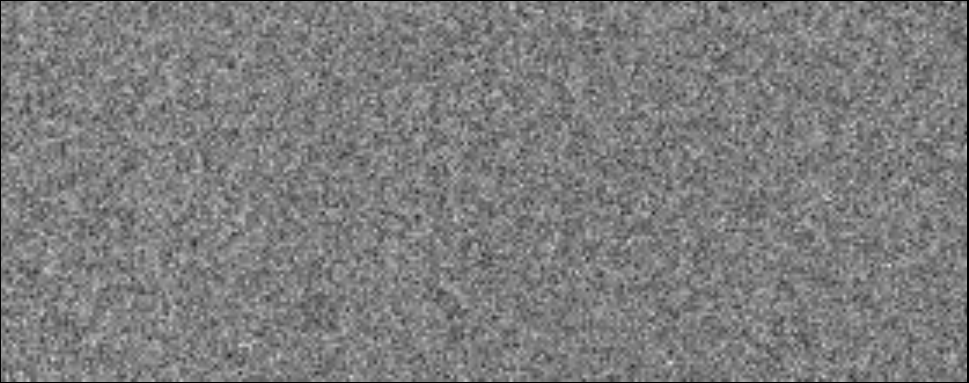}
\includegraphics[width=2.8cm]{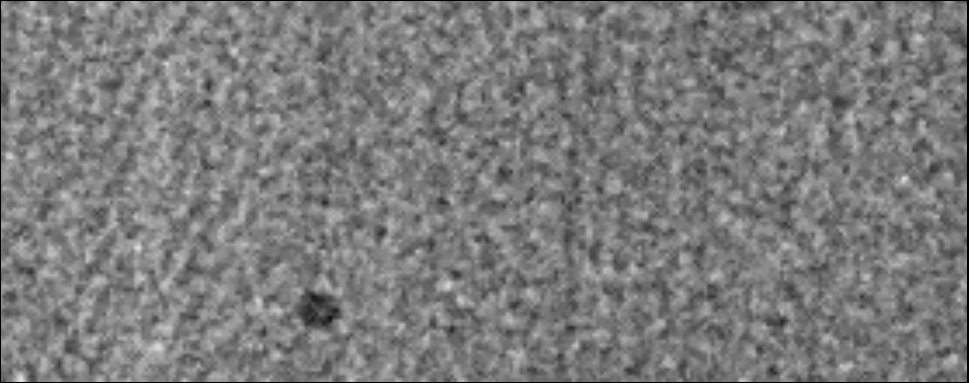}
\includegraphics[width=2.8cm]{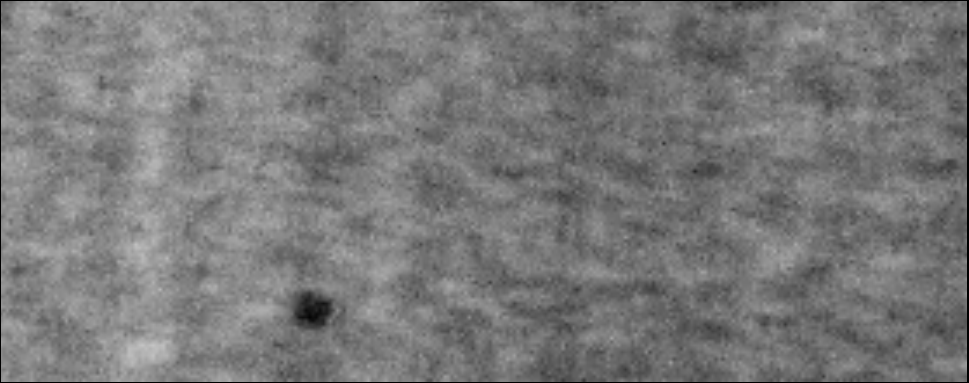}
\end{minipage}
}

\vspace{1mm}

\centerline{
\begin{minipage}{\linewidth}
  \begin{overpic}[width=4.25cm]{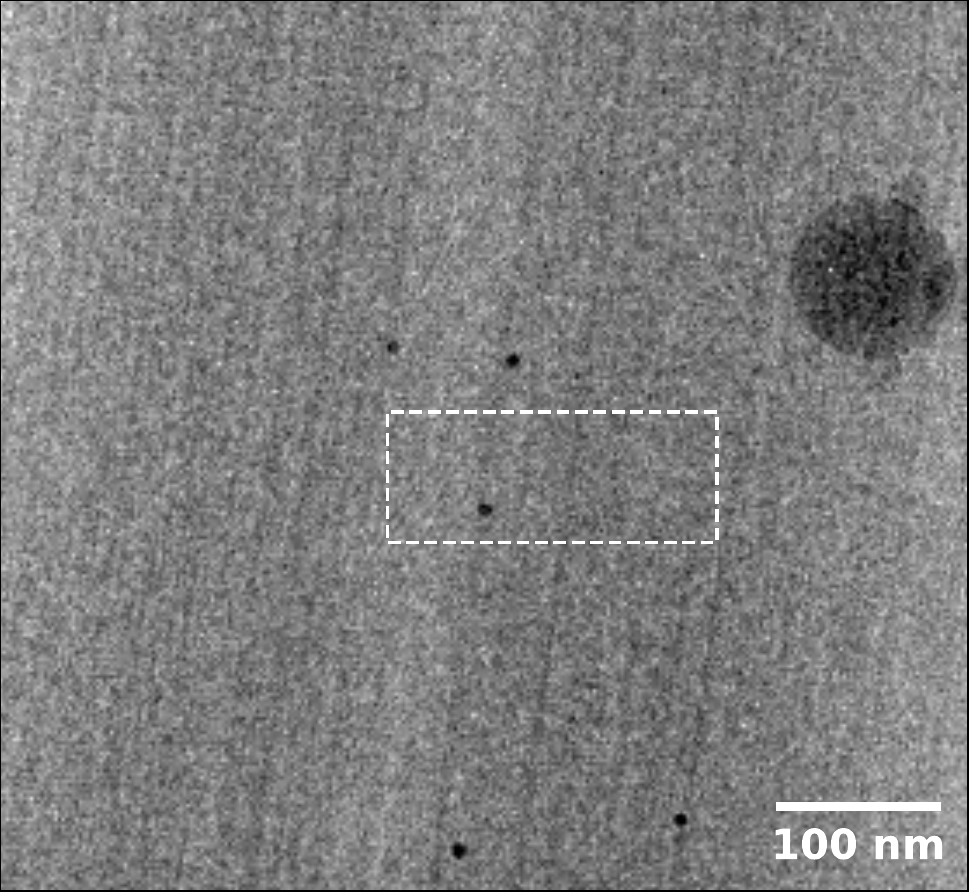}
    \put (0.5,2) {\figNum{white}{(d) P2P-\textit{ip}}}
  \end{overpic}
  \begin{overpic}[width=4.25cm]{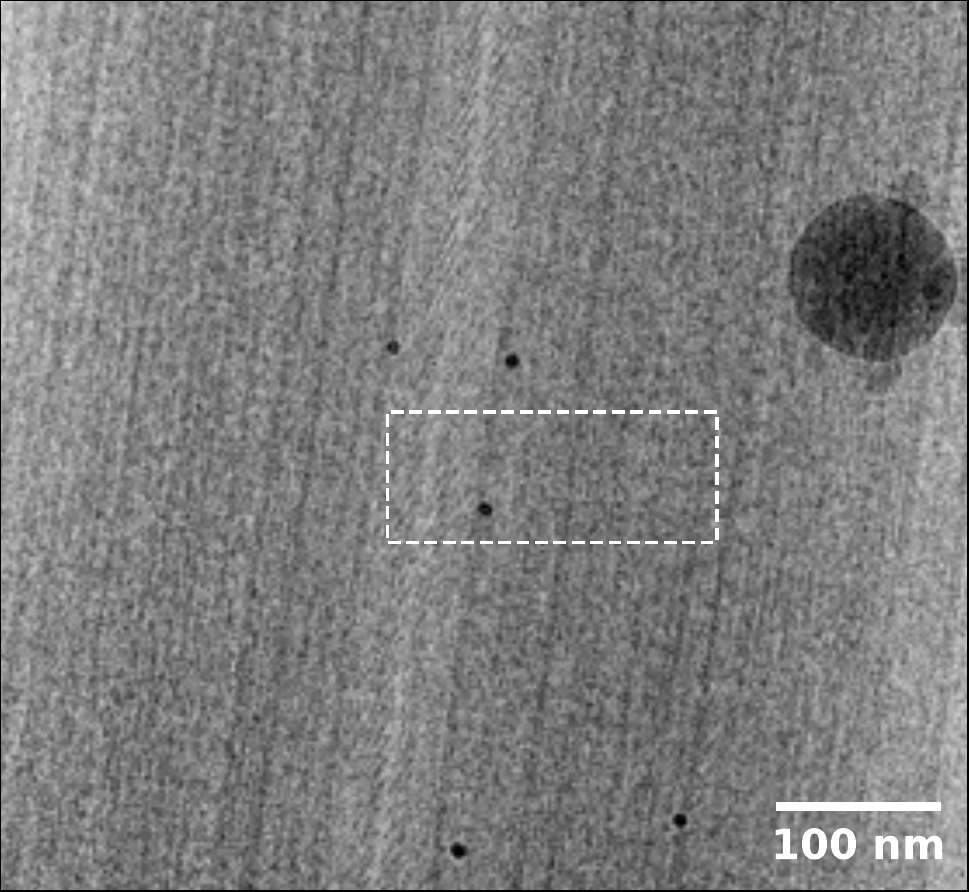}
  \put (0.5,2) {\figNum{white}{(e) P2P-\textit{df}}}
  \end{overpic}
\end{minipage}
}

\centerline{
\begin{minipage}{\linewidth}
\centering
\includegraphics[width=4.25cm]{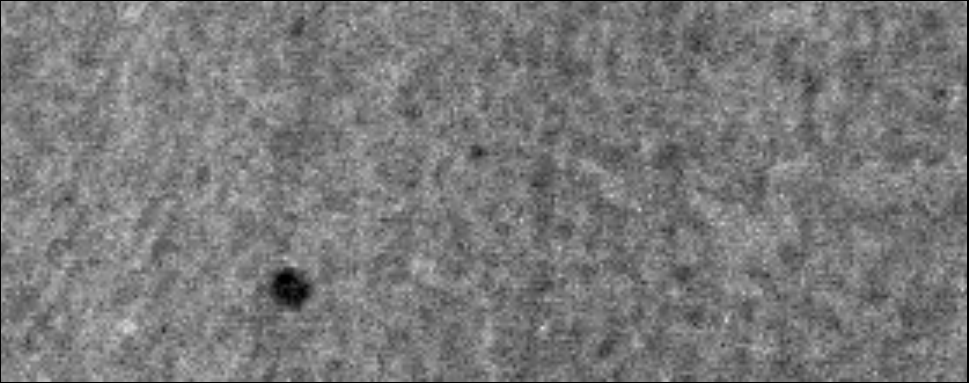}
\includegraphics[width=4.25cm]{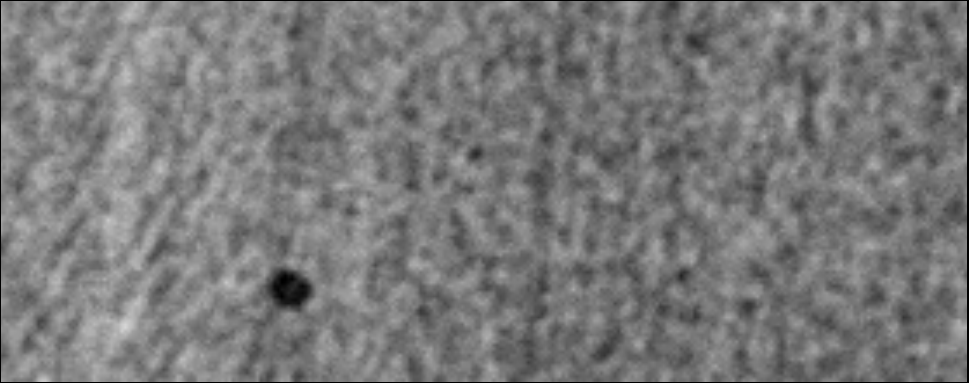}
\end{minipage}
}

\caption{Cryo-\CARE results on a 2D cryo-TEM projection. 
Subfigures and insets show: 
raw input data~(a), 
median filtered restoration baseline~(b), 
Cryo-\CARE results when trained on tomographic tilt-angle pairs~(c), 
on acquired image pairs~(d), and on dose-fractionated movie frames~(e).}
  \label{fig:p2p}
\end{figure}
}

\newcommand\figTtoT{
\begin{figure}[tb]
\begin{minipage}[b]{\linewidth}
\hspace{5.1mm}
 \begin{minipage}{\linewidth}
    \begin{overpic}[width=3.88cm]{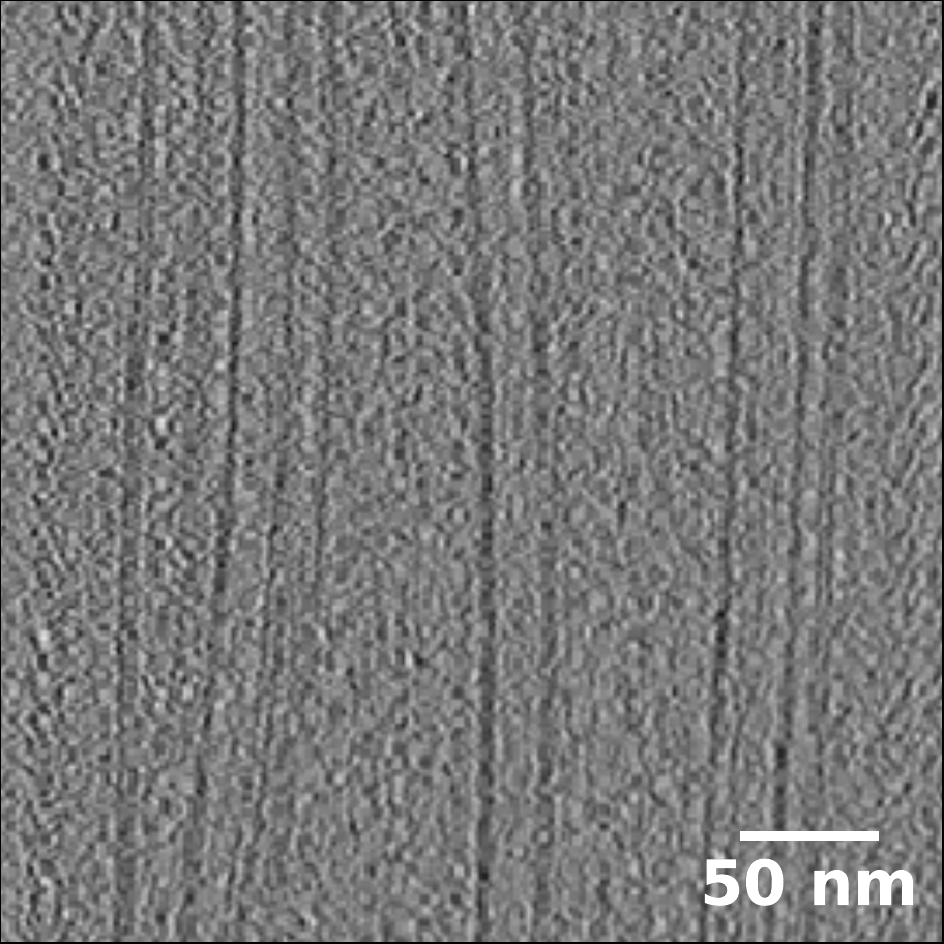}
        \put (0.5,2) {\figNum{white}{(a)}}
    \end{overpic}
    \begin{overpic}[width=3.88cm]{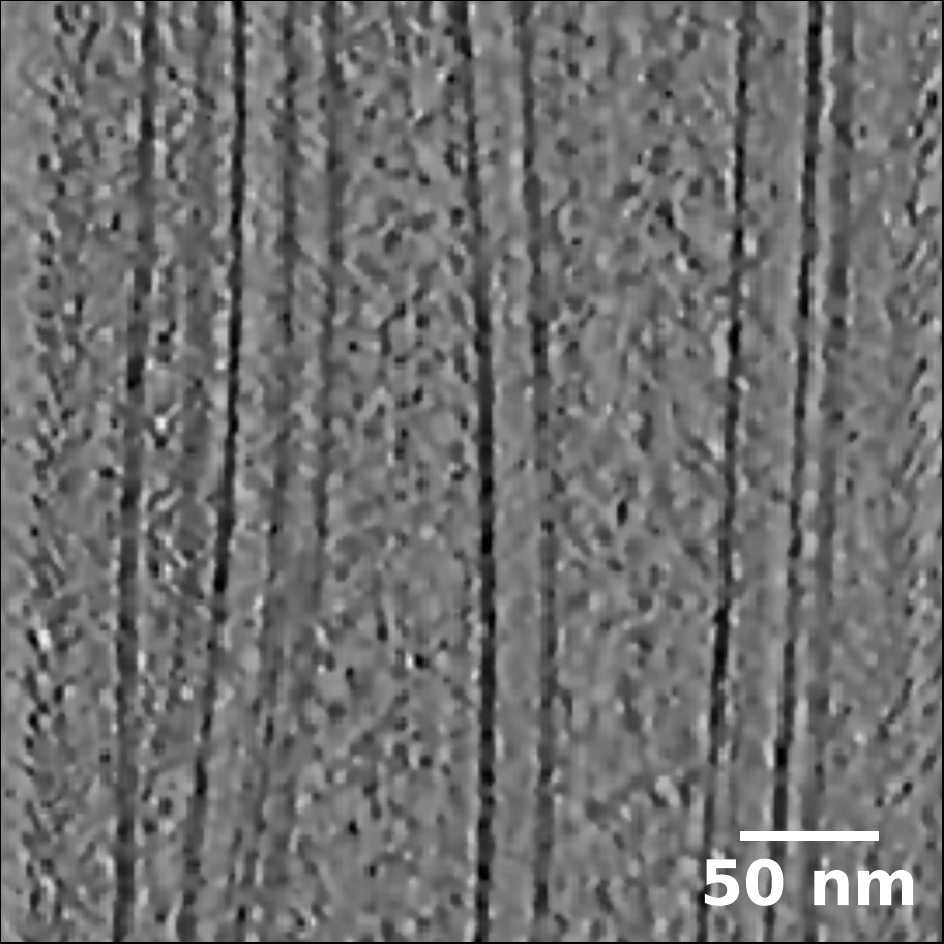}
        \put (0.5,2) {\figNum{white}{(b)}}
    \end{overpic} 
 \end{minipage}
 
 \vspace{0.75mm}
 
\hspace{5.1mm}
 \begin{minipage}{\linewidth}
  \begin{overpic}[width=3.88cm]{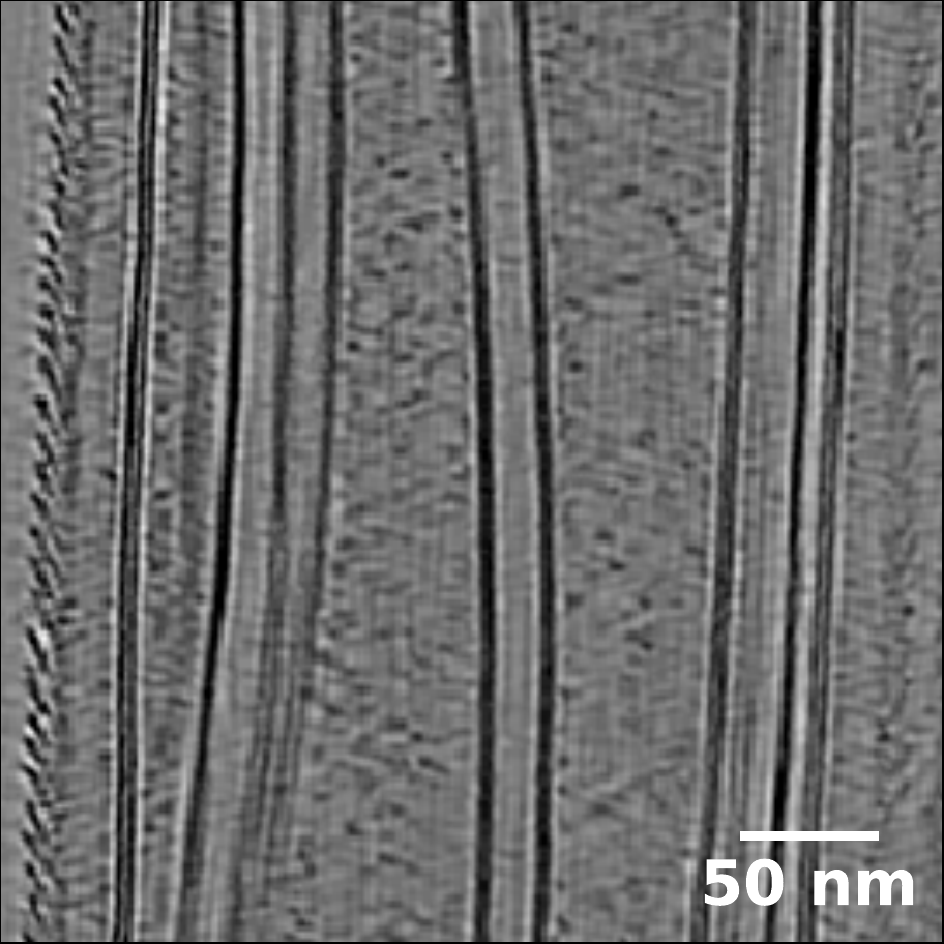}
    \put (0.5,2) {\figNum{white}{(c) T2T-\textit{eoa}}}
  \end{overpic}
  \begin{overpic}[width=3.88cm]{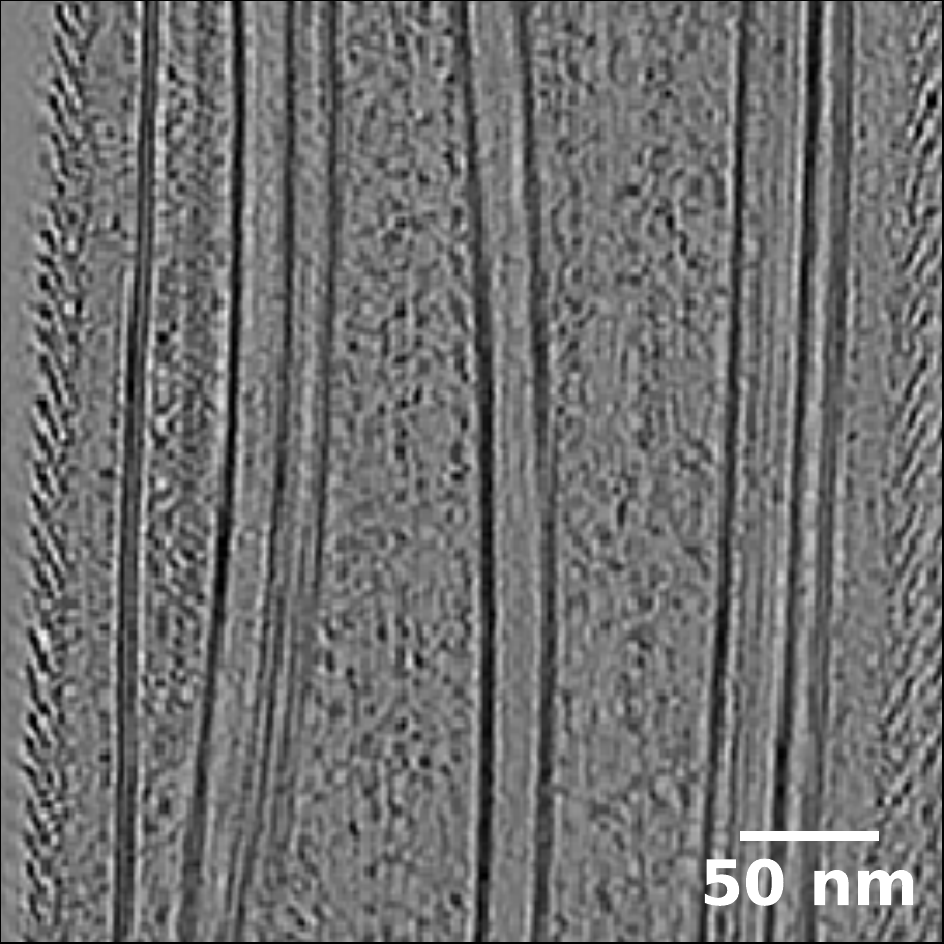}
    \put (0.5,2) {\figNum{white}{(d) T2T-\textit{df}}}
  \end{overpic}
 \end{minipage}
\end{minipage}
\begin{minipage}[b]{1.0\linewidth}
\centering
 \centerline{\includegraphics[width=8.5cm]{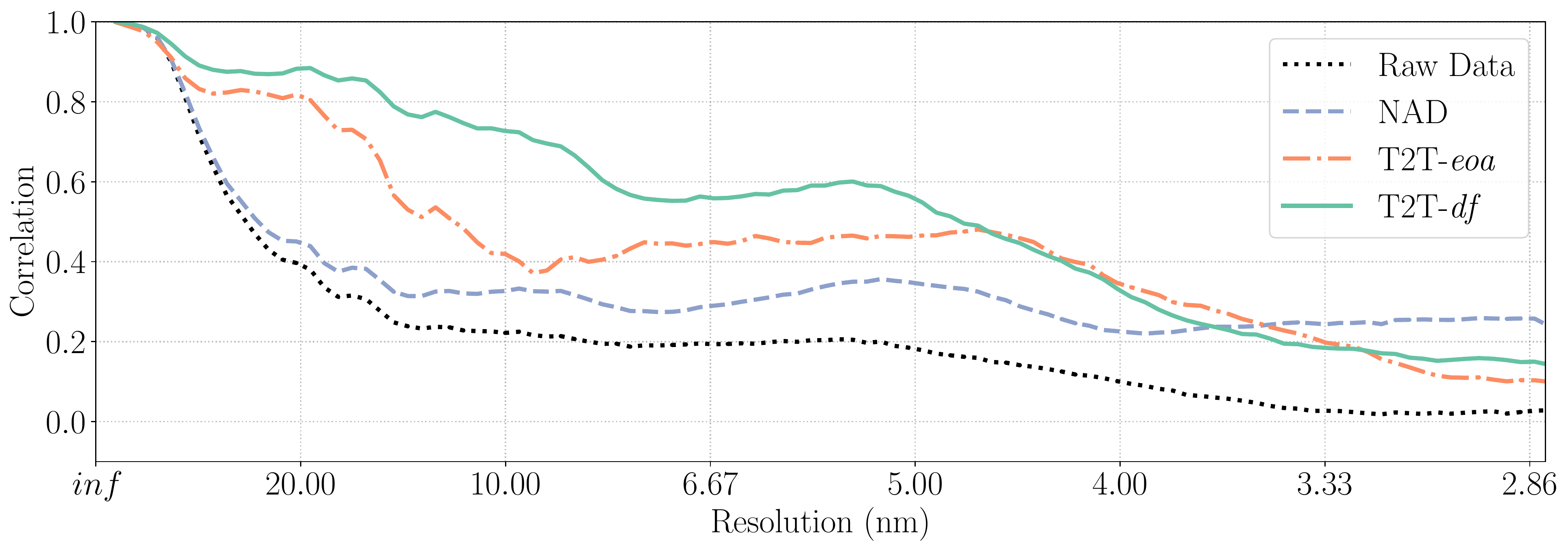}}
\end{minipage}
\caption{Cryo-\CARE results on a 3D cryo-TEM tomogram.
Subfigures show: 
a section through the raw tomogram~(a),
the non-linear anisotropic diffusion filtered baseline~(b),
cryo-\CARE results when trained via our proposed \TtoT-\textit{eoa}~(c) and \TtoT-\textit{df}~(d). 
The graph shows the Fourier shell correlation (FSC) curves of the raw tomogram, the baseline, and our proposed methods.}
\label{fig:t2t}
\end{figure}
}

\newcommand\figSegmentation{
\begin{figure}[ht]
\hspace{0.67cm}\begin{minipage}[b]{7.61cm} 
 \begin{overpic}[width=7.58cm]{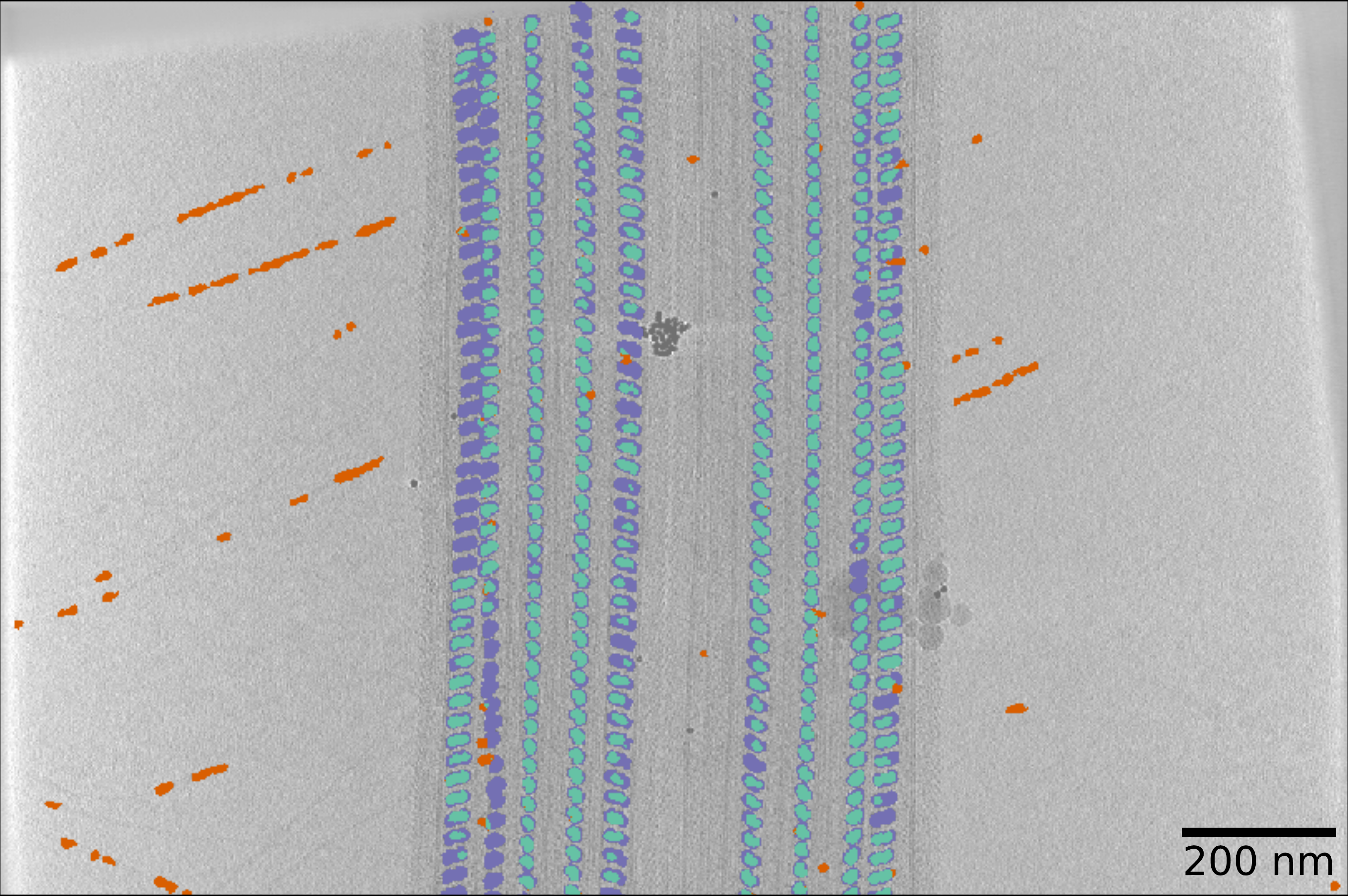}
    \put (0.5,1) {\figNum{black}{(a)}}
 \end{overpic}
 
 \vspace{0.5mm}
 
 \begin{overpic}[width=7.58cm]{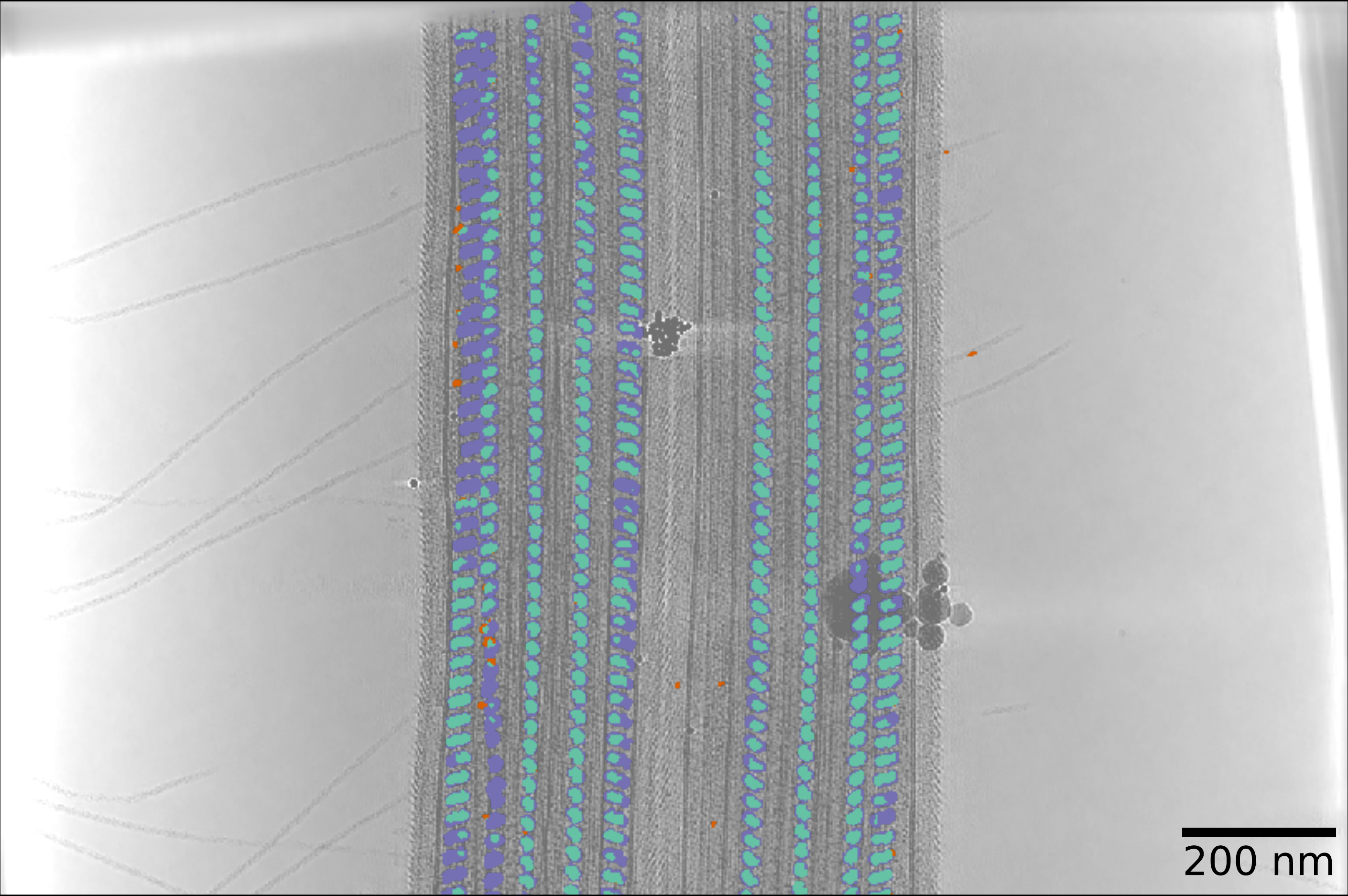}
    \put (0.5,1) {\figNum{black}{(b)}}
 \end{overpic}
\end{minipage}

\vspace{.5mm}

\begin{minipage}[b]{1.0\linewidth}
 \centering
 \centerline{\includegraphics[width=8.5cm]{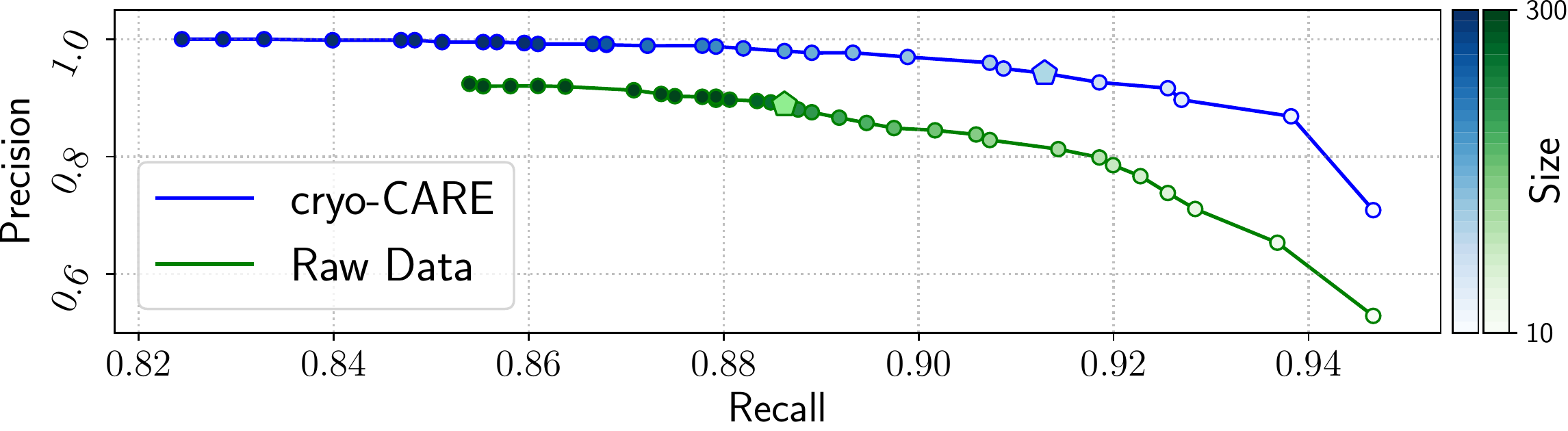}}
  \vspace{-2mm}
\end{minipage}

\caption{
Automated downstream analysis on 
raw data~(a) and a \TtoT-\textit{df} restored tomogram~(b).
Ground-truth voxels are shown in violet, true-positives in turquoise, and false-positives in orange. 
Precision-recall plots on increasing segment size threshold (see main text) are shown below. 
The pentagons correspond to sub-figures (a) and (b).}
\label{fig:segmentation}
\end{figure}
}

\newcommand\figArtifacts{
\begin{figure}[htb]
\centerline{
\begin{minipage}{\linewidth}
\begin{overpic}[width=.49\linewidth]{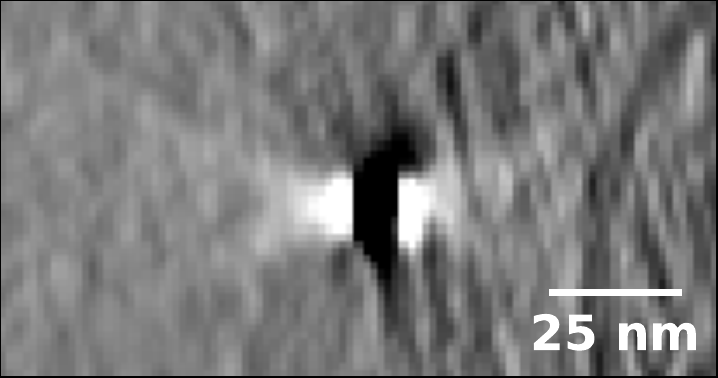}
 \put (0.5,2) {\figNum{white}{(a)}}
\end{overpic}
\begin{overpic}[ width=.49\linewidth]{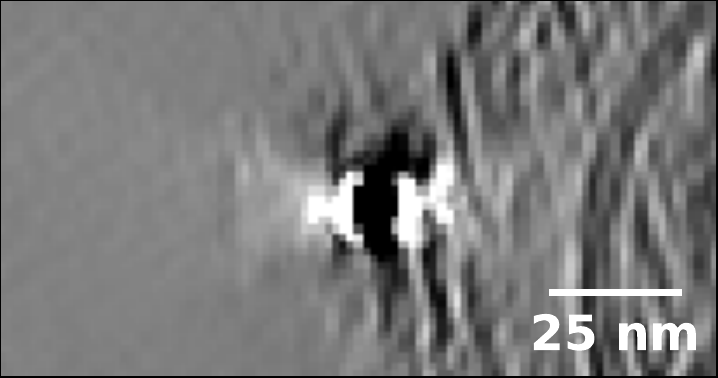}
 \put (0.5,2) {\figNum{white}{(b)}}
\end{overpic}
\end{minipage}
}

\caption{Tomogram reconstruction artifacts. 
Tomograms reconstructed from \PtoP restored tilt-angles lead to strong missing-wedge artifacts~(a). 
This problem is reduced using our proposed \TtoT training scheme~(b).}
\label{fig:artifacts}
\end{figure}
}

\newcommand\figEMPIAR{
\begin{figure}[b]
\centering
\centerline{
\begin{minipage}{\linewidth}
\begin{overpic}[width=2.82cm]{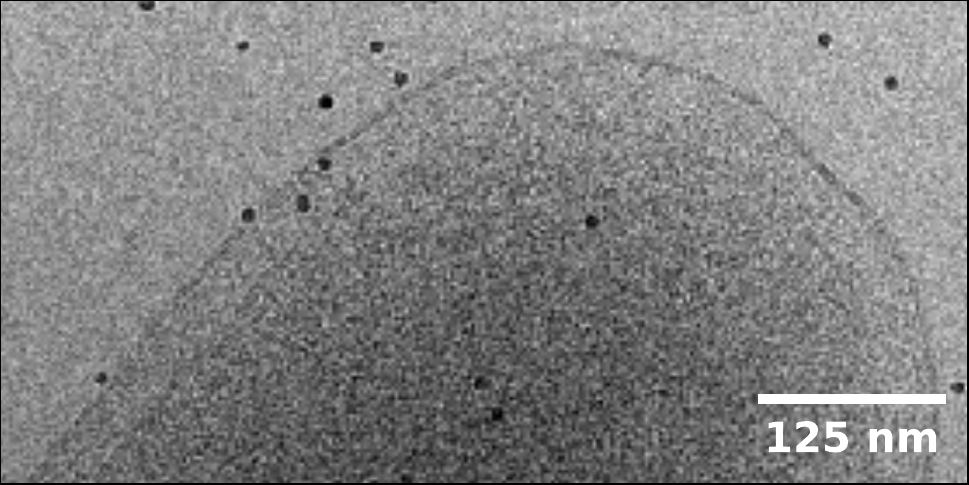}
 \put (0.5,2) {\figNum{white}{(a)}}
\end{overpic}
\begin{overpic}[width=2.82cm]{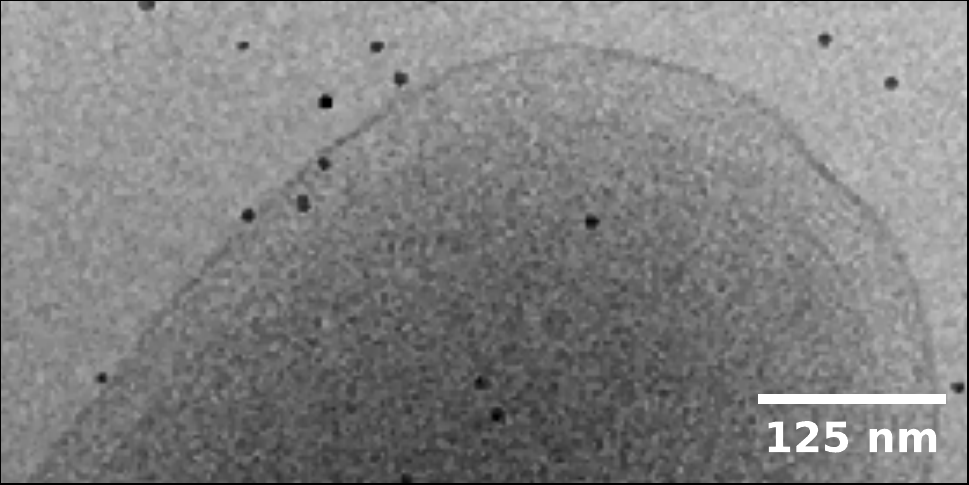}
 \put (0.5,2) {\figNum{white}{(b)}}
\end{overpic}
\begin{overpic}[width=2.82cm]{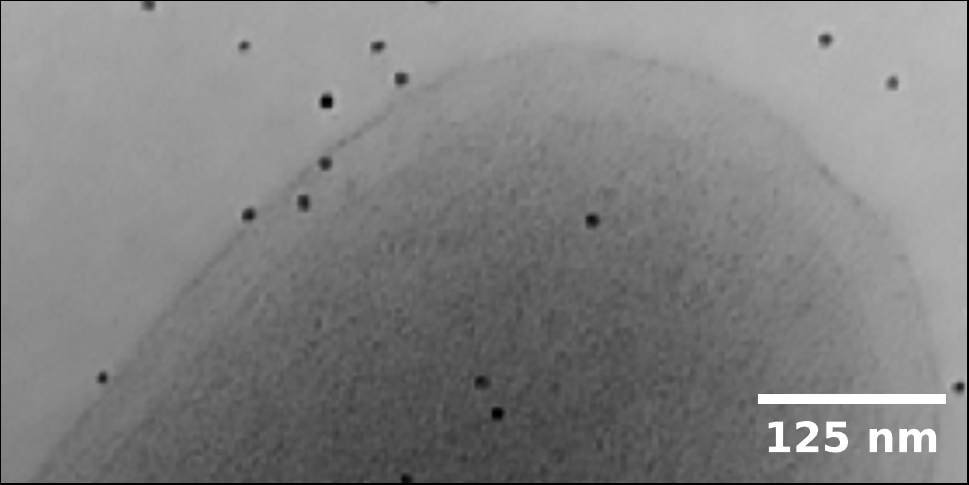}
 \put (0.5,2) {\figNum{white}{(c)}}
\end{overpic}
\end{minipage}
}

\vspace{1mm}

\centerline{
\begin{minipage}{\linewidth}
\begin{overpic}[width=2.82cm]{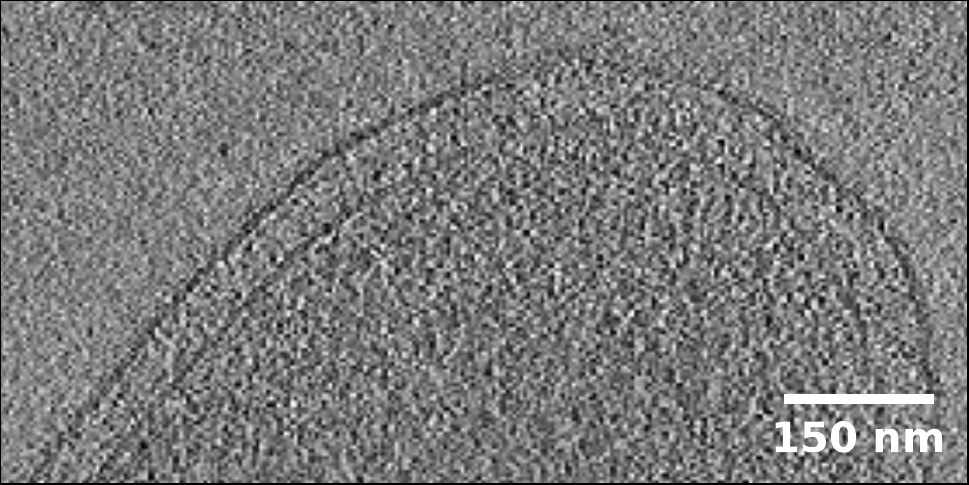}
 \put (0.5,2) {\figNum{white}{(d)}}
\end{overpic}
\begin{overpic}[width=2.82cm]{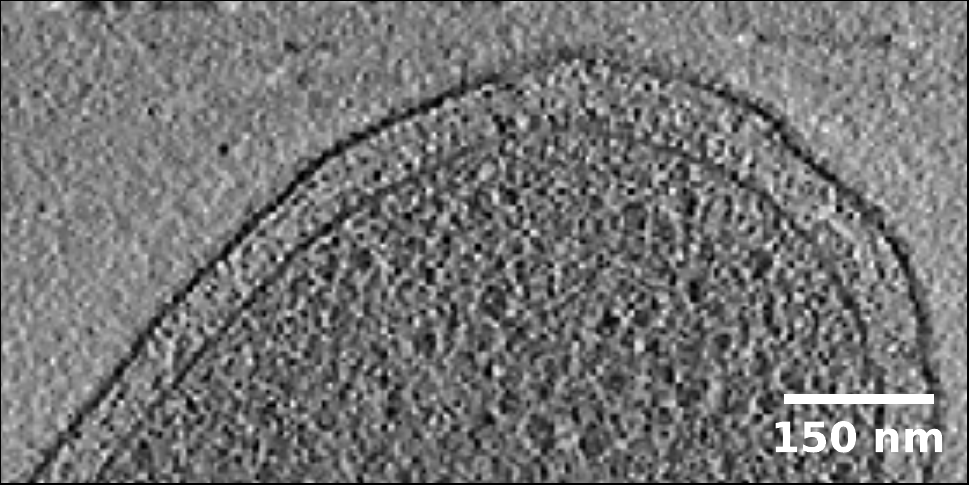}
 \put (0.5,2) {\figNum{white}{(e)}}
\end{overpic}
\begin{overpic}[width=2.82cm]{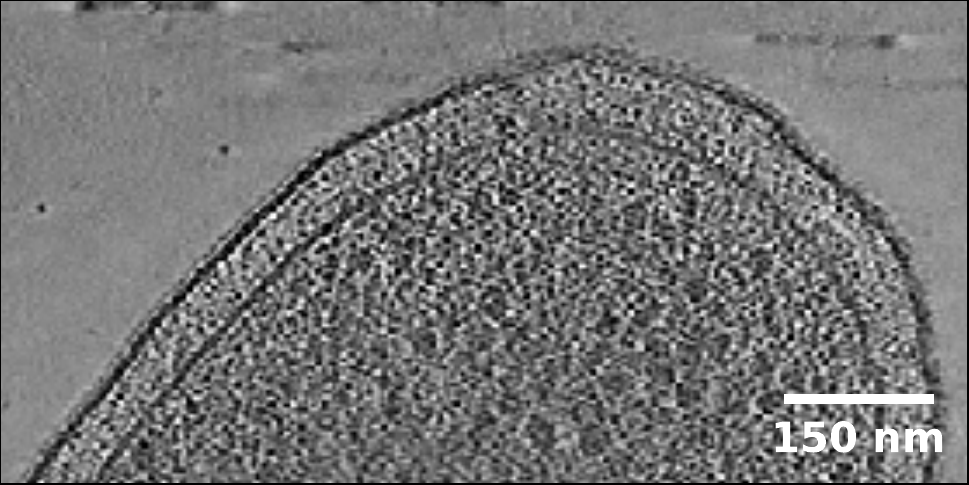}
 \put (0.5,2) {\figNum{white}{(f)}}
\end{overpic}
\end{minipage}
}

\caption{Cryo-\CARE restoration on the publicly available \EMPIAR dataset. 
(a)~Raw projection (single tilt angle). 
(b)~Median filtered baseline. 
(c)~Our \PtoP-tap results. 
(d)~Raw tomogram. 
(e)~NAD filtered baseline. 
(d)~Our even-odd \TtoT restoration.}
\label{fig:EMPIAR}
\end{figure}
}

\begin{abstract}
Multiple approaches to use deep learning for image restoration have recently been proposed.
Training such approaches requires well registered pairs of high and low quality images.
While this is easily achievable for many imaging modalities, \eg fluorescence light microscopy, for others it is not. 
Cryo-transmission electron microscopy (cryo-TEM) could profoundly benefit from improved denoising methods, unfortunately it is one of the latter.
Here we show how recent advances in network training for image restoration tasks, \ie denoising, can be applied to cryo-TEM data.
We describe our proposed method and show how it can be applied to single cryo-TEM projections and whole cryo-tomographic image volumes.
Our proposed restoration method dramatically increases contrast in cryo-TEM images, which improves the interpretability of the acquired data.
Furthermore we show that automated downstream processing on restored image data, demonstrated on a dense segmentation task, leads to improved results.
\end{abstract}
%
\section{Introduction}
\label{sec:intro}
\vspace{-2mm}

\begin{textblock*}{86.2mm}(19mm,260mm)
$^*$ This work has been submitted to the IEEE for possible publication. Copyright may be transferred without notice, after which this version may no longer be accessible.
\end{textblock*}

Modern cryo-transmission electron microscopy (cryo-TEM) enables the observation of biological structures in their native state at high resolution. 
In order to prevent sample destruction during image acquisition, the total electron dose needs to be restricted~\cite{knapek1980beam}.
This restriction results in noisy, low contrast acquisitions.
In practice, electron microscopists typically acquire defocused images to trade resolution for increased contrast.
Hence, the existence of better performing image restoration methods would enable image acquisitions at low electron dose with reduced defocus and therefore also at elevated resolution.

For fluorescence microscopy data, deep learning can be used for content-aware image restoration (\CARE)~\cite{weigert2017content}. 
Data for training \CARE networks requires adequately imaged or synthetically generated pairs at low and high quality.
The ideas presented in \cite{weigert2017content} do not translate to cryo-TEM data, where the before mentioned electron dose prevents the acquisition of non-noisy ground truth images.

Here we present cryo-\CARE, a way to train content-aware restoration networks for cryo-TEM data. 
Cryo-\CARE can be trained by using registered pairs of noisy images, an idea that was recently introduced in the context of real-world RGB and MRI images~\cite{lehtinen2018noise2noise}. 
More concretely, we show how single TEM projections and whole tomographic volumes can be denoised using a strong, learned, and content-aware prior.

We compare our results to simple baseline methods such as median-filtering or NAD~\cite{frangakis2001noise}.
Despite their simplicity, these methods are widely used by cryo-TEM experts to improve the interpretability of their data.
Furthermore we show that automated downstream processing on restored image data leads to significantly improved results.

\section{Approach and Methods}
\label{sec:methods}
\vspace{-2mm}
Since cryo-TEM acquisitions suffer from very low signal-to-noise ratio (SNR) due to the limited electron dose, recording high SNR ground truth images is not possible.
Recently it was demonstrated that image denoising networks can be trained without having high SNR images available. Instead, only well registered pairs of low SNR images are required~\cite{lehtinen2018noise2noise}. 
Hence, for cryo-TEM data a combination of \CARE and the ideas from~\cite{lehtinen2018noise2noise} enable us, for the first time, to train cryo-\CARE networks.

\compactsubsub{Network Architecture and Training Procedure}{
For all experiments we used a U-Net~\cite{ronneberger2015u} of depth two, a convolution kernel size of three, and a linear activation function at the last layer.
Moreover we used a per-pixel mean squared error loss. 
In all experiments we keep 10\% of extracted training patches (see below) as validation set.
Note that we use the open-source \CARE framework~\cite{weigert2017content} for all experiments.
}

\subsection{Restoring Single cryo-TEM Projections (\PtoP)}
\label{sec:p2p}
\figPtoP

Here we describe three ways to train cryo-\CARE networks on adequately prepared pairs of cryo-TEM projections.
Hence, we call this approach \ProjectionProjection or \PtoP.

\compactsubsub{\textbf{\PtoP-ip}, training using acquired image pairs}{
The most straight forward way to combining \CARE~\cite{weigert2017content} and \NoiseNoise~\cite{lehtinen2018noise2noise} is to acquire pairs of images for which the noise is independent.

To this end, we acquired such image pairs on a 300~kV Thermo Fisher cryo-TEM Titan Halo that is equipped with a K2 direct electron detector from Gatan. 
More precisely, we acquired images of \textit{Chlamydomonas reinhardtii} cilia in dose-fractionation mode (movie mode)~\cite{li2013electron}, splitted the frames in two halves, and averaged them without additional alignment, giving us the equivalent of two independently acquired images at half the available electron dose each\footnote{Note that each image in such a pair has an even lower SNR due to the halved electron dose.}.
From such pairs of images we extracted $1000$ randomly selected patch-pairs of size $128\times 128$ which are used to train a \CARE network in the \NoiseNoise regime. 
After training, we use the trained network to restore all image pairs and retrieve the final result by per-pixel averaging the two individual restorations (see Fig.~\ref{fig:p2p}(d)). 
}

\compactsubsub{\textbf{\PtoP-tap}, using tomographic tilt-angle pairs}{
For readily acquired, not dose-fractionated data, the previously described scheme cannot be applied.
Archived data for which only single acquisition exist can therefore not be used for training cryo-\CARE networks.
For existing tilt-series, acquired for tomographic reconstruction, we asked ourselves if pairs of neighboring tilt-angles could be used for training. 
We used IMOD~\cite{kremer1996computer} to align and register all acquired tilt-angles. 
As before, training was performed on $1000$ randomly selected patch pairs of size $128\times 128$ taken from adjacent tilt-angle projections.
Final restorations are retrieved by applying the trained network to both tilt-angles individually (see Fig.~\ref{fig:p2p}(c)).
}

\compactsubsub{\textbf{\PtoP-df}, using dose-fractionated movie frames}{
Since our data was acquired on a Gatan K2 direct detector, we were able to go an additional step further. 
Instead of using two acquired images, as described initially, we can leverage the fact to have many more frames acquired.
As it is usually done during dose-fractionation, we can additionally correct for motion-blur of the sample by registering the individual frames using MotionCor2~\cite{Zheng061960}.
We then sum all even and odd frames to retrieve two images with independent noise. 
This interleaved frame-splitting is advantageous because induced beam damage will be equally shared in both independent images. 
Again we trained on $1000$ randomly selected patch pairs of size $128\times 128$, and created the final restored projection by applying the network to both images followed by per-pixel averaging (see Fig.~\ref{fig:p2p}(e)).
}

\subsection{Restoration of cryo-TEM Tomograms (\TtoT)}
\label{sec:t2t}
\figTtoT

Also here we present three ways to restore cryo-TEM tomograms.
One obvious way is to first use \PtoP cryo-\CARE to restore all tilt-angles individually and then reconstruct a tomogram from them.
All tomographic reconstructions were performed with ETOMO, which is part of IMOD~\cite{kremer1996computer}.
We will later see that this leads to clearly visible artifacts that can be mitigated by first reconstructing two tomograms, each from half the available data, and then train a 3D cryo-\CARE network in a \NoiseNoise regime.
We call these approaches collectively \TomoTomo, or \TtoT in short.

\compactsubsub{\textbf{\TtoT-\textit{eoa}}, using even-odd acquisitions}{
This protocol is designed to work for conventionally acquired tilt-series, when no direct detector is available.
Here we split all tilted projections in two sets based on their acquisition number. 
From all tilt angles with an even/odd acquisition number, we reconstructed two data-independent tomograms and used those to train a 3D cryo-\CARE network on $1200$ randomly selected 3D sub-volumes of size $64\times 64\times 64$.
We created the final restored tomogram by applying the trained network to both tomograms followed by per-pixel averaging (see Fig.~\ref{fig:t2t}(d)).
}

\compactsubsub{\textbf{\TtoT-\textit{df}}, using dose-fractionated movie frames}{
In case the available data was acquired in dose-fractionation mode (movie mode), we propose a slightly different protocol.
For each tilt-angle, similar to our \PtoP approach on dose-fractionated data, we can equally split all frames, align and sum them. The two sets of independent tilt-angle projections can then be used to reconstruct two independent tomograms.
We trained as before and created the final restored tomogram by applying the trained network to both tomograms followed by per-pixel averaging.
The advantage of this approach is that the angular sampling for both tomograms is denser and consistent, hence leading to better results  (see Fig.~\ref{fig:t2t}(d)). 
}

\subsection{Automated Downstream Analysis}
\label{sec:autoanalysis}
We train a dense segmentation and detection workflow for \textit{Chlamydomonas reinhardtii} outer dynein arm (ODA). 
Also here we use a U-Net~\cite{ronneberger2015u} trained on manually generated and with PEET~\cite{Nicastro2006,Heumann2011} refined ground truth. 
The predicted segmentation is then normalized and Otsu thresholded~\cite{otsu1979threshold}. 
Each connected component is then filtered according to its size in voxels. 
Remaining components are treated as one detected ODA. 
Since we had only one hand annotated tomogram, we split annotations in train and test set, counting $383$ and $712$, respectively.
We did not use data augmentation.

\section{Results}
\label{sec:results}
\vspace{-2mm}
We evaluated all variations of our proposed methods on two datasets. 
The first one, called \TOMO, was acquired by ourselves on a 300~kV Thermo Fisher cryo-TEM Titan Halo with a Gatan K2 direct electron detector. 
This enables us to test cryo-\CARE variations that require image acquisitions in dose-fractionation mode (movie mode).
The second dataset, \EMPIAR\footnote{http://dx.doi.org/10.6019/EMPIAR-10110}, is publicly available via the EMPIAR database~\cite{iudin2016empiar} and consists of a complete tomographic series of tilted projections. 
We did not find publicly available raw dose-fractionated data but will make our \TOMO data available upon publication.

P2P results are computed on unbinned data and T2T results are computed on six times binned data.

\figEMPIAR

\subsection{\PtoP Restoration Experiments}
We first tested \PtoP-\textit{tap}, which is applicable to both available datasets (see Fig.~\ref{fig:p2p}(c) and Fig.~\ref{fig:EMPIAR}(c)). 
Using adjacent tilt-angles pairs during training leads to restored images that appear blurry. 
This is due to the relative displacement of individual structures in the imaged volume when projected at adjacent tilt-angles.

To circumvent this problem, we train cryo-\CARE networks on specifically acquired pairs of images (\PtoP-\textit{ip}) and dose-fractionated movie frames (\PtoP-\textit{df}).
Restorations using these networks are shown in Fig.~\ref{fig:p2p}(d,e).

\compactsubsub{Tomographic reconstruction from \PtoP restored tilt-angles}{
A canonical idea to reconstruct denoised tomograms is to use restored \PtoP-\textit{df} tilt angles (like in Fig.~\ref{fig:p2p}(e)). 
This does, unfortunately, amplify the missing wedge artifacts at high-gradient locations (see Fig.~\ref{fig:artifacts}).
Since neural networks are complex non-linear filters and tilt-angle reconstructions are performed independently, the predicted intensities for a given structure is not necessarily consistent across restored tilt-angles.
These inconsistent amplitudes are likely the reason for the amplification of the observed missing wedge artifacts.
Nevertheless, this problem can be addressed with the \TtoT network training regimes described in Section~\ref{sec:methods}. 
}

\figArtifacts

\subsection{\TtoT Restoration Experiments}
In Fig.~\ref{fig:t2t} we compare \TtoT even-odd acquisitions and \TtoT dose-fractionated to the reconstructed raw tomogram and a non-linear anisotropic diffusion filtered (NAD) baseline. 
For a quantitative comparison we computed the Fourier shell correlation (FSC) for all reconstructions. 
We computed the FSC for the raw data directly from the two tomograms we created for \TtoT dose-fractionated training. 
For the NAD filtered tomogram and our \TtoT dose-fractionated results, we computed the FSC on the filtered and restored data, respectively. 
For the \TtoT even-odd acquisitions we split, as commonly done, the available tilt-angles in order to compute the FSC.
The slices we show in Fig.~\ref{fig:t2t}(a-d) allow a qualitative assessment of the power of cryo-CARE, while the plotted FSC curves quantify the gained image quality.
Still, the question remains if cryo-\CARE is improving the automated analysis of acquired cryo-TEM data.

\subsection{Cryo-\CARE Facilitates Automated Analysis}
In order to test if the improved SNR and contrast in restored cryo-TEM tomograms benefits automated downstream analysis, we implemented a segmentation and detection workflow (described in Section~\ref{sec:autoanalysis}). 
Fig.~\ref{fig:segmentation} shows results of the exact same automated analysis performed on raw tomograms and on cryo-\CARE restored data.
In Fig.~\ref{fig:segmentation} we can appreciate a significant increase in precision and recall when data is first restored and then analyzed. 

\figSegmentation

\section{Discussion}
\label{sec:discussion}
\vspace{-2mm}
In this publication we show how content-aware image restoration can successfully be applied to cryo-TEM data. 
EM experts are currently using relatively simple filtering techniques, \ie NAD, before manually investigating acquired data.
Cryo-\CARE, as we have shown, leads to highly contrasted and well resolved 2D and 3D data.
Our experiments also show that \PtoP reconstructions are not ideal for tomographic reconstructions. 
Nevertheless, with \TtoT we can offer a simple and powerful tool for content-aware tomographic restorations.
We therefore believe that cryo-\CARE will facilitate manual data browsing, a step that can hardly be underestimated when many and/or large volumes have to be browsed for regions of interest.

Additionally we showed that cryo-\CARE restorations can lead to highly improved automated analysis results.
An essential feature of cryo-\CARE is that training data can be generated by the microscope itself and does not require tedious human labeling.
While end-to-end pipelines on raw data might need huge amounts of labeled data to also co-learn to restore the noisy data, cryo-\CARE helps to uncouple these two tasks -- a preprocessing step that does not need human labels and a analysis stage that is likely to require lesser amounts of training data.

We are confident that cryo-\CARE will rapidly find application in the cryo-EM field.
It improves data-browsing, creates well contrasted, high SNR images for improved visualization of single projections/tomograms, and improves the performance of automated analysis pipelines, hence it enables to work efficiently on much larger bodies of data.

\vfill

\bibliographystyle{IEEEbib}
\bibliography{strings,refs}

\end{document}